\pdfoutput=1
\documentclass[journal]{IEEEtran}
\usepackage{cite}
\usepackage{amsmath}
\usepackage{amssymb}
\usepackage{graphicx}
\usepackage{multirow}
\usepackage{adjustbox}
\usepackage{array}
\usepackage{booktabs}
\usepackage{xcolor}
\usepackage{placeins}
\usepackage{eso-pic}
\usepackage{hyperref}
\hypersetup{pdftitle={RGB\textendash{}Event Object Detection Across Operating Frequencies via Reliability-Supervised Routing and Asymmetric-Resolution Fusion}, pdfauthor={Peijun Shi, Jiannan Zhao, Xiujin Zhu, Zhenguo Lao, Chee-Onn Chow, Wei Ru Wong}}

\definecolor{evfill}{RGB}{226,238,251}
\definecolor{evdraw}{RGB}{42,120,214}
\definecolor{rgbfill}{RGB}{253,234,222}
\definecolor{rgbdraw}{RGB}{214,90,40}
\definecolor{newfill}{RGB}{222,244,233}
\definecolor{newdraw}{RGB}{20,130,90}

\newcommand{\sg}{\mathrm{sg}}

\newcommand{\biophoto}[1]{\IfFileExists{#1}%
  {\includegraphics[width=1in,height=1.25in,clip,keepaspectratio]{#1}}%
  {\fcolorbox{red}{white}{\parbox[c][\dimexpr1.25in-2\fboxsep-2\fboxrule\relax][c]{\dimexpr1in-2\fboxsep-2\fboxrule\relax}%
    {\centering\color{red}\footnotesize photo\\missing}}}}

\begin{document}
\AddToShipoutPictureFG*{\AtPageLowerLeft{\raisebox{0.3in}{\makebox[\paperwidth]{\parbox{\textwidth}{\centering\footnotesize This work has been submitted to the IEEE for possible publication. Copyright may be transferred without notice, after which this version may no longer be accessible.}}}}}
\bstctlcite{IEEEexample:BSTcontrol}%

\title{RGB--Event Object Detection Across Operating Frequencies via Reliability-Supervised Routing and Asymmetric-Resolution Fusion}

\author{Peijun~Shi, Jiannan~Zhao, Xiujin~Zhu, Zhenguo~Lao, Chee-Onn~Chow,~\IEEEmembership{Senior~Member,~IEEE}, Wei~Ru~Wong%
\thanks{P. Shi, X. Zhu, Z. Lao, C.-O. Chow, and W. R. Wong are with the Department of Electrical Engineering, Faculty of Engineering, Universiti Malaya, 50603 Lembah Pantai, Kuala Lumpur, Malaysia.}%
\thanks{J. Zhao is with the School of Electrical Engineering, Guangxi University, Nanning 530004, China, and also with the Computational Intelligence Laboratory, School of Computer Science, University of Lincoln, Lincoln LN6 7TS, U.K.}%
\thanks{Corresponding author: Chee-Onn Chow (e-mail: cochow@um.edu.my).}}

\markboth{Submitted to IEEE Transactions on Image Processing}%
{Shi \MakeLowercase{\textit{et al.}}: RGB--Event Object Detection Across Operating Frequencies}

\maketitle

\begin{abstract}
Autonomous vehicles must detect objects faster than a frame camera delivers images. Event cameras record brightness changes with microsecond resolution, so RGB--event detectors can decide between frames, at an operating frequency, the rate of decisions, far above the frame rate. As this rate rises, each decision receives a sparser event window, and the reliability of the events relative to the frame varies over the scene and over time. Existing detectors weight the sensors per sample and encode the frame at the reduced event resolution. Reliability-supervised routing assigns a fusion weight to every location and supervises it with targets from the losses of two single-modality classifiers used only in training, moving loss-based supervision of fusion weights from whole samples to individual locations. Asymmetric-resolution fusion lets the frame carry the decision where the events fail: it encodes the frame at native resolution and pools every feature level onto the event grid, reusing all pretrained weights. On the DSEC-Det dataset, routing halves the accuracy loss from 20 to 180 Hz, from 6.3 to 3.1 mAP, and lowers it in every seed. The resulting detector, ERA-Det, sets a new state of the art at 63.2 mAP, 3.7 mAP above FlexEvent, leads at every operating frequency, and at 180 Hz is more accurate than FlexEvent at 20 Hz. On two DAVIS346 datasets, the router lowers the frame weight at night and in low light and raises it when the events vanish. ERA-Det upgrades released FlexEvent weights in 5,000 iterations on one GPU.
\end{abstract}


\section{Introduction}
\label{sec:intro}

\IEEEPARstart{A}{utonomous} vehicles must detect objects with a latency that matches the speed of the traffic around them. A frame camera sampling at 20~Hz delivers one image every 50~ms, during which a car driving at 50~km/h moves 69~cm. Event cameras report brightness changes asynchronously with microsecond resolution and a dynamic range of about 140~dB~\cite{gallego2022survey}, so a detector that combines frames and events can issue decisions between frames. Gehrig and Scaramuzza showed that this combination reaches the latency of a 5,000~fps camera at the bandwidth of a 45~fps camera~\cite{gehrig2024dagr}. The rate of these decisions is the operating frequency, distinct from the frame rate, which stays at 20~Hz: at 180~Hz, nine decisions share one frame, and the same car moves 7.7~cm between two decisions. RGB--event detection at high operating frequencies is therefore a direct route to safer perception, and FlexEvent~\cite{lu2025flexevent} formalized it as detection across operating frequencies.

The accuracy of current RGB--event detectors does not hold across this range. Raising the operating frequency shortens the event window that each decision receives, from 50~ms at 20~Hz to 5.6~ms at 180~Hz, and the event representation becomes sparse. The released FlexEvent model loses 13.1~mAP between 20~Hz and 180~Hz on the DSEC-Det dataset~\cite{gehrig2024dagr}, and its variant trained with pseudo-labels at intermediate timestamps still loses 6.4~mAP. When the events become sparse, the RGB frame is the dense source of evidence, and two properties of existing fusion detectors limit how well this evidence is used. First, the fusion weight is decided once per sample~\cite{lu2025flexevent} or per expert~\cite{bao2026hyperfeod}, whereas the reliability of the events relative to the frame changes from one location to the next: a static car is sharp in the frame and produces few events, and a fast pedestrian produces dense events. Second, the RGB branch is encoded at the reduced resolution of the event representation, so a sensor that records $640\times430$ pixels of the scene contributes only a quarter of them, and small objects such as pedestrians lose most of their appearance.

Both properties treat the events as uniformly reliable, whereas their reliability relative to the frame varies over the scene and over time. The principle of this paper is that the fusion should track, location by location and decision by decision, which sensor is more reliable; robustness across operating frequencies is its measurable instance, as a shorter event window lowers the reliability of the events at a fixed frame. Dynamic fusion methods for classification derive one weight per modality and sample from the confidence or the losses of unimodal predictors~\cite{han2022dynamics,zhang2023qmf}. Reliability-supervised routing realizes this principle: it predicts a fusion weight at every location and trains it with a target derived from the losses of two single-modality classifiers, which measure at each object location which sensor explains the object better. Routing can move the decision to the frame only as far as the frame provides evidence, and asymmetric-resolution fusion supplies this evidence: since the event encoder runs at every decision and the RGB encoder only once per frame, it keeps the events at the resolution of the base detector, encodes the frame at its native resolution and maps every RGB feature level onto the grid of the event stage it is fused with, through a depthwise pooling initialized as exact averaging, so all pretrained weights of the base detector load without modification. We refer to the resulting detector as ERA-Det, for event--RGB asymmetric detector.

ERA-Det sets a new state of the art on DSEC-Det. It reaches 63.2~mAP on the test split, 3.7~mAP above FlexEvent under an identical protocol, and each of three training seeds exceeds FlexEvent. It is the most accurate detector at every operating frequency from 20~Hz to 180~Hz and without events. At 180~Hz it reaches 61.3~mAP, 8.7~mAP above the pseudo-label variant of FlexEvent that was trained for this regime, and keeps 95\% of its 20~Hz accuracy, so it decides nine times as often as FlexEvent at 20~Hz and is still more accurate. The advantage holds with the seven annotated classes of DSEC-Det, where routing lowers the accuracy loss from 20~Hz to 180~Hz from 7.7 to 4.9~mAP. Both components also transfer to two datasets recorded with DAVIS346 sensors, the PKU-DDD17-Car dataset~\cite{li2019jdf} and the PKU-DAVIS-SOD dataset~\cite{li2023sodformer}, whose frames have the resolution of the events. Asymmetric-resolution fusion raises FlexEvent by 0.9~mAP on both and achieves the best reported accuracy on PKU-DDD17-Car, 53.0~mAP and 91.2~AP$_{50}$, and the best reported AP$_{50}$ on PKU-DAVIS-SOD, 58.5. The router lowers the weight of the frame at night and in low light in every seed and raises it when the events vanish, and without events routing adds 1.5~mAP on PKU-DAVIS-SOD.

The contributions of this paper are as follows.
\begin{itemize}
\item The principle that RGB--event fusion should follow the reliability of the events relative to the frame at every location, realized as reliability-supervised routing: a fusion weight per location, supervised by the losses of two single-modality classifiers used only in training. It carries loss-based supervision of fusion weights~\cite{han2022dynamics,zhang2023qmf} from whole samples to individual locations and, to our knowledge, makes ERA-Det the first RGB--event detector whose fusion weights are supervised at every location. Routing halves the accuracy loss between 20~Hz and 180~Hz, from 6.3 to 3.1~mAP, lowers it in every training seed and with seven classes, and adds 4.0~mAP without events.
\item Asymmetric-resolution fusion, which gives the frame the evidence that routing relies on by setting the resolution of the RGB branch independently of the event branch through a depthwise pooling layer initialized as exact averaging, reusing every pretrained weight of the base detector. It raises the Global accuracy on DSEC-Det by 3.9~mAP and the accuracy at 180~Hz by 11.1~mAP. On two DAVIS346 datasets, it achieves the best reported accuracy on PKU-DDD17-Car and the best reported AP$_{50}$ on PKU-DAVIS-SOD. Its additional computation lies in the RGB encoder, which runs once per frame.
\item ERA-Det, a frequency-robust RGB--event detector that sets a new state of the art on DSEC-Det and exceeds FlexEvent at every operating frequency from 20~Hz to 180~Hz and without events, with two and with seven classes. It upgrades released weights in 5,000 iterations on a single GPU, and a controlled analysis over three seeds isolates the contribution of each component.
\end{itemize}

The code will be released at \url{https://github.com/Peijun1122/ERA-Det}. Section~\ref{sec:related} reviews related work, Section~\ref{sec:method} presents ERA-Det, Section~\ref{sec:experiments} reports the experiments, and Section~\ref{sec:conclusion} concludes the paper.

\section{Related Work}
\label{sec:related}

\subsection{Event-Based Object Detection}
Learning-based event detectors convert a window of events into a dense tensor and process it with a recurrent or attention-based network. RED~\cite{perot2020red} combined convolutional features with ConvLSTM memory~\cite{shi2015convlstm}, and RVT~\cite{gehrig2023rvt} introduced multi-axis attention with recurrent memory at every stage. SAST~\cite{peng2024sast} sparsifies windows and tokens according to the scene, SSM~\cite{zubic2024ssm} replaces the recurrent layers with state-space models whose time scales adapt to the window length, and LEOD~\cite{wu2024leod} trains event detectors with limited labels. Static objects produce few events and must be retained in the recurrent memory, which motivates frames as a complementary source. ERA-Det adopts the RVT encoder for events and focuses on the fusion of the two sensors.

\subsection{RGB--Event Fusion for Detection}
RGB--event detectors differ mainly in how and where the two modalities are combined. RENet~\cite{zhou2023renet} fuses multi-scale event frames with RGB features through bidirectional attention, CAFR~\cite{cao2024cafr} refines both modalities from coarse to fine with cross-modal interaction, and SFNet~\cite{liu2024sfnet} injects speed-invariant event frames into the image branch. SODFormer~\cite{li2023sodformer} treats detection as streaming sequence prediction with asynchronous attention, and DAGr~\cite{gehrig2024dagr} pairs a convolutional image branch with an asynchronous graph network that updates detections between frames. FAOD~\cite{faod} aligns RGB features to events with deformable convolution under frame-rate mismatch. Recent designs make the fusion adaptive: FlexEvent~\cite{lu2025flexevent} weights the two modalities with a learned per-sample gate, FocusMamba~\cite{yang2025focusmamba} selects tokens guided by event activity, Hyper-FEOD~\cite{bao2026hyperfeod} routes each sample to a subset of experts, and illumination-adaptive and knowledge-aware fusion schemes adapt the combination to the scene~\cite{shi2026review}. In these detectors the RGB branch operates at the resolution of the event representation, and the fusion weights are either shared by all locations of a sample or of an expert or learned without a signal on which sensor is reliable. ERA-Det decouples the two resolutions and supervises a weight at every location.

\subsection{Detection Across Operating Frequencies}
A detector that runs between frames receives event windows shorter than those it was trained on. SSM~\cite{zubic2024ssm} addresses this for event-only detection with learnable time scales, and DAGr~\cite{gehrig2024dagr} reuses the image features of the latest frame and updates detections from new events. FlexEvent~\cite{lu2025flexevent} established the evaluation of RGB--event detection from 20~Hz to 180~Hz on DSEC-Det and introduced FlexTune, which generates pseudo-labels at intermediate timestamps to train the detector at high frequencies. FlexEvent exceeds DAGr and the event-only detectors by more than 15~mAP on the DSEC-Det test split~\cite{lu2025flexevent} and is therefore the reference for this problem. ERA-Det exceeds FlexTune at high frequencies without pseudo-labels, because its fusion relies on the frame where the events become sparse.

\subsection{Reliability-Aware Fusion and Routing}
Weighting each source by an estimate of its quality is a central principle of multisensor fusion~\cite{zhang2024lowquality}. Illumination-aware detectors weight color and thermal branches by the predicted lighting~\cite{guan2019illumination,zhou2020mbnet}, CMX~\cite{zhang2023cmx} rectifies each modality with weights predicted from the other, and EGM-Det~\cite{egmdet} gates RGB--infrared features with an entropy prior. Mixture-of-experts models make the fusion conditional through a learned router~\cite{mees2016choosing,cao2023moefusion,shazeer2017moe,fedus2022switch}, and their routers are trained by the task loss together with load-balancing terms. Jointly trained multimodal networks, however, favor the modality that is easiest to fit~\cite{wang2020makes,peng2022ogm}, so a router trained by the fused detection loss alone receives no signal about which sensor is correct at a given location. Dynamic fusion methods for classification supervise the weights directly: Multimodal Dynamics~\cite{han2022dynamics} learns the confidence of each modality from the true-class probability of a unimodal classifier, and QMF~\cite{zhang2023qmf} shows that fusion weights which decrease with the unimodal losses tighten the generalization bound of the fused model. In both, the supervised weight is assigned per modality to a whole sample. ERA-Det is, to our knowledge, the first RGB--event detector to supervise its fusion weights at every location of a detection feature pyramid: the losses of two single-modality classifiers define a per-location target that measures the reliability of the events relative to the frame, and a contrastive term~\cite{oord2018representation,rusak2024infonce,tian2020cmc} aligns the object-level representations of the two sensors.

\begin{figure*}[!t]
\centering
\includegraphics[width=\textwidth]{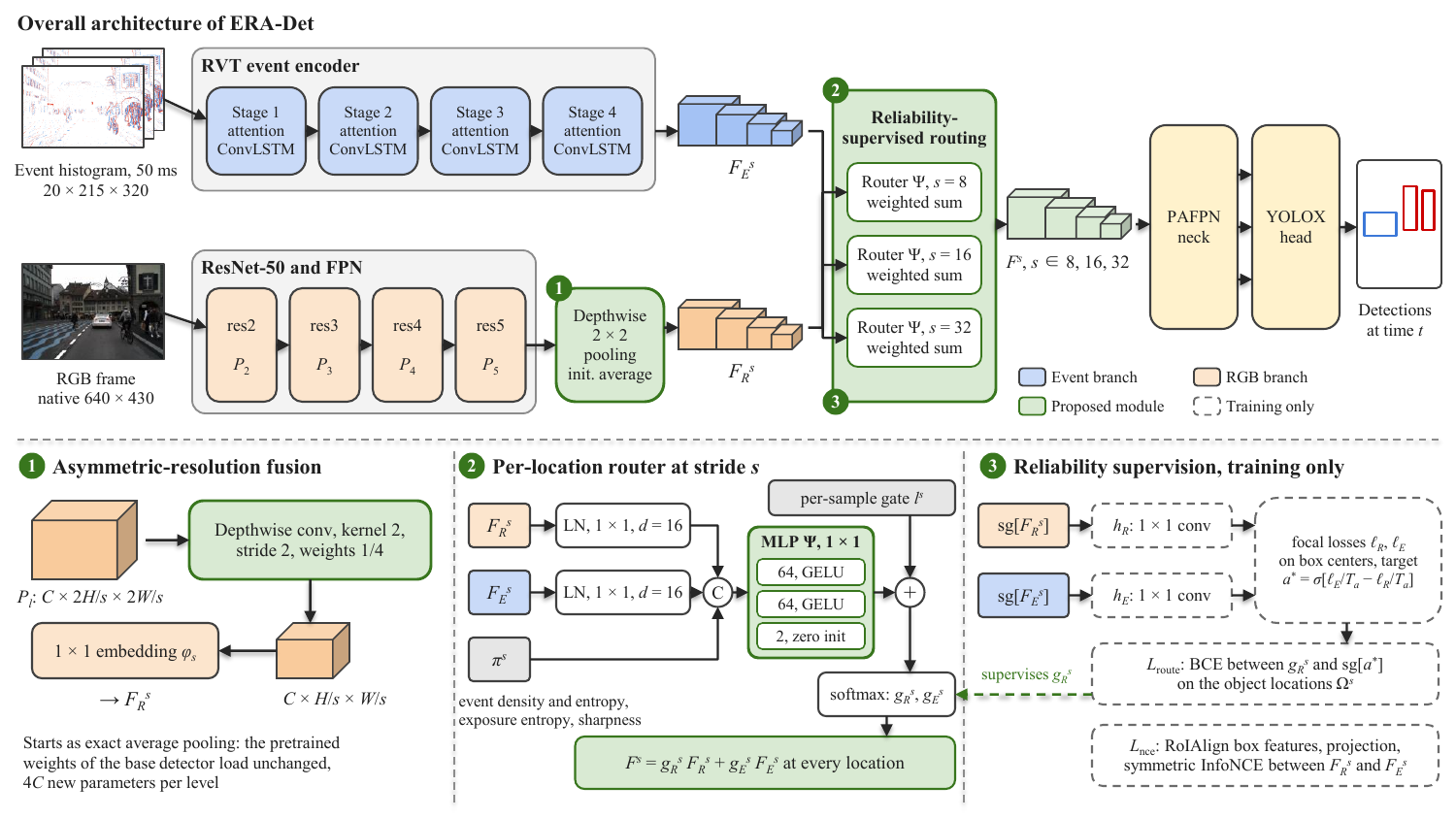}
\caption{Architecture of ERA-Det. Top: the event histogram and the native-resolution RGB frame are encoded by an RVT encoder and a ResNet-50 with a feature pyramid, asymmetric-resolution fusion maps every RGB level onto the event grid, and reliability-supervised routing fuses the two pyramids at strides 8, 16 and 32 before the PAFPN neck and the YOLOX head. Panel 1: asymmetric-resolution fusion, a depthwise pooling initialized as exact averaging. Panel 2: the per-location router, which corrects the per-sample gate of the base detector at every location. Panel 3: the training-only reliability supervision, whose routing target is computed from two single-modality classifiers and supervises the RGB weight of the router. The inputs are a frame and the following 50~ms of events of the DSEC-Det sequence thun\_02\_a.}
\label{fig:framework}
\end{figure*}

\section{Method}
\label{sec:method}

\subsection{Problem Formulation and Overview}
\label{sec:overview}
An event camera produces a stream $\mathcal{E}=\{e_i=(x_i,y_i,t_i,p_i)\}$ of brightness changes with pixel location $(x_i,y_i)$, timestamp $t_i$ and polarity $p_i$. A frame camera registered to the event camera delivers frames $I(t_f)$ at a period $T_f$. At an evaluation time $t\in[t_f,t_f+T_f]$, the detector receives the latest frame and the events in $(t_f,t]$ and predicts the objects present at $t$. The operating frequency is $1/(t-t_f)$: at 20~Hz on DSEC-Det the detector receives 50~ms of events, at 180~Hz it receives 5.6~ms, and at $t=t_f$ it receives no events. The operating frequency is thus a decision rate, while the frame period stays at $T_f$, 50~ms on DSEC-Det. A detector is frequency-robust when its accuracy is stable over this range.

ERA-Det builds on the architecture of FlexEvent~\cite{lu2025flexevent}, shown in Fig.~\ref{fig:framework}. The events are converted into stacked histograms with 10 temporal bins and two polarities on a $320\times215$ grid, for a primary window $A$ and a shorter window $B$ inside it, and an RVT encoder~\cite{gehrig2023rvt} produces event features $F^s_E$ at strides $s\in\{4,8,16,32\}$ for each window. A ResNet-50 with a feature pyramid~\cite{he2016resnet,lin2017fpn} encodes the frame, a $1\times1$ embedding maps every pyramid level to the channel width of the corresponding event stage, and a fusion module combines the two modalities at every stride. A PAFPN neck~\cite{liu2018panet} and a YOLOX head~\cite{ge2021yolox} decode the fused features at strides 8, 16 and 32. ERA-Det changes two parts of this pipeline: the resolution of the RGB branch, by asymmetric-resolution fusion, abbreviated ARF and described in Section~\ref{sec:arf}, and the fusion at strides 8, 16 and 32, by reliability-supervised routing, abbreviated RSR and described in Section~\ref{sec:rsr}. ERA-Det is FlexEvent with both components, and FlexEvent + ARF is ERA-Det without RSR.

\subsection{Asymmetric-Resolution Fusion}
\label{sec:arf}
The DSEC frames are registered to the event camera at $640\times480$ pixels, and the detector uses the top $430$ rows. FlexEvent encodes the events at half resolution, as RVT does on the 1~Mpx dataset~\cite{gehrig2023rvt}, and resizes the frame crop to the same $320\times215$ grid so that the RGB features at every stride match the event grid. The RGB encoder needs to run only once per frame, yet the resize discards three quarters of the pixels before it sees them, and the loss is largest for small objects, whose detection accuracy depends on the scale at which the encoder sees them~\cite{singh2018snip}. Asymmetric-resolution fusion sets the two resolutions separately and moves the change of resolution from the image to the feature pyramid: the events keep the grid of the base detector, and the frame is encoded at $k=2$ times this resolution, its native $640\times430$ on DSEC-Det. On sensors whose frame has the resolution of the events, such as the DAVIS346 sensors of Sections~\ref{sec:ddd17} and~\ref{sec:sod}, the frame is upsampled by $k$, which presents every object to the pretrained RGB encoder at a larger scale.

Let $P_l\in\mathbb{R}^{C\times kH/2^l\times kW/2^l}$ be the pyramid level of stride $2^l$ on the native frame, where $H\times W$ is the event grid and $C=256$. The RGB feature fused with the event stage of stride $s=2^l$ is
\begin{equation}
F^s_R=\phi_s\big(D^s_k(P_l)\big),\qquad D^s_k(P_l)\in\mathbb{R}^{C\times H/s\times W/s},
\label{eq:arf}
\end{equation}
where $\phi_s$ is the $1\times1$ embedding of the base detector and $D^s_k$ is a depthwise convolution with kernel size $k$ and stride $k$. Every weight of $D^s_k$ is initialized to $1/k^2$, so that at the start of training $D^s_k$ computes exact $k\times k$ average pooling.

This construction has three properties that make it a practical upgrade of an existing detector. First, each pooled level keeps the pyramid level, the channel semantics and the spatial size of the input that the fusion module was trained on, so all weights of the base detector load unchanged and only $4C$ parameters per level are new. Taking the next coarser level $P_{l+1}$ of the native frame, which already has the size of the event grid, would instead feed every fusion module a level it was not trained with and leave the finest level $P_2$ of the native frame unused. Second, each pooled feature summarizes $k^2$ positions that were computed from the full-resolution frame, so the information that the resize would have removed reaches the fusion module. Third, the pooling is learned, so every channel can weight its four inputs according to the detail it carries. The event branch, the fusion module and the detection head operate at the original resolution, so the additional cost is confined to the RGB encoder. During training, the activations of the RGB encoder are recomputed in the backward pass instead of being stored~\cite{chen2016checkpoint}; ERA-Det therefore trains on sequences of 11 windows with 16 sequences per batch within 34.3~GB of GPU memory.

Augmentation keeps the two resolutions aligned: flips and rotations act on each tensor about its own center, and for zoom-in and zoom-out the crop window on the frame is exactly $k$ times the window sampled on the event grid, in corner and size.

\subsection{Reliability-Supervised Routing}
\label{sec:rsr}
The fusion module of FlexEvent computes, for each sample and each window, two gate logits $\mathbf{l}^s\in\mathbb{R}^2$ from the concatenated features and forms $g^s_RF^s_R+g^s_EF^s_E$ with weights shared by all locations. Reliability-supervised routing makes the weights depend on the location and supervises them with a measured target.

\textit{Per-location router.} At stride $s\in\{8,16,32\}$ and location $\mathbf{x}$, the router receives projections of both features and a set of observation statistics,
\begin{equation}
\mathbf{z}^s(\mathbf{x})=\big[\mathbf{W}_R\,\mathrm{LN}(F^s_R),\ \mathbf{W}_E\,\mathrm{LN}(F^s_E),\ \boldsymbol{\pi}^s\big](\mathbf{x}),
\label{eq:router_in}
\end{equation}
where $\mathrm{LN}$ normalizes over channels, $\mathbf{W}_R$ and $\mathbf{W}_E$ project to $d=16$ channels, and $\boldsymbol{\pi}^s$ holds four statistics computed without learned parameters: the density and the time--polarity entropy of the events and the exposure entropy and the sharpness of the frame in the patch of $\mathbf{x}$. The statistics give the router direct access to the observation conditions; Section~\ref{sec:ablation} shows that the supervised router extracts the same information from the features, so they can be omitted. A three-layer perceptron $\Psi$ with 64 hidden channels, applied at every location, predicts a correction of the per-sample logits, and the fusion weights and the fused feature are
\begin{equation}
\begin{aligned}
\big(g^s_R(\mathbf{x}),\,g^s_E(\mathbf{x})\big)&=\mathrm{softmax}\big(\mathbf{l}^s+\Psi(\mathbf{z}^s(\mathbf{x}))\big),\\
F^s(\mathbf{x})&=g^s_R(\mathbf{x})\,F^s_R(\mathbf{x})+g^s_E(\mathbf{x})\,F^s_E(\mathbf{x}).
\end{aligned}
\label{eq:route}
\end{equation}
The last layer of $\Psi$ is initialized to zero, so that the router starts from the per-sample gating of the base detector. The fused maps of the two event windows are averaged, as in FlexEvent. The router adds 0.05~M parameters over the three strides and does not change the feature dimensions that the neck receives.

\textit{Routing target.} The weight $g^s_R(\mathbf{x})$ should be large where the RGB feature identifies the object better than the event feature. We measure this with two auxiliary classifiers $h^s_R$ and $h^s_E$, implemented as $1\times1$ convolutions on the stop-gradient features $\sg(F^s_R)$ and $\sg(F^s_E)$. Each classifier predicts the class map of the objects whose central region covers a location, and its per-location focal loss~\cite{lin2017focal} is $\ell^s_M(\mathbf{x})$ for $M\in\{R,E\}$. Because the classifiers read stopped features, they assess the two modalities without changing the encoders. The routing target and the routing loss are
\begin{equation}
a^\ast(\mathbf{x})=\sigma\!\left(\frac{\ell^s_E(\mathbf{x})-\ell^s_R(\mathbf{x})}{T_a}\right),
\label{eq:target}
\end{equation}
\begin{equation}
\mathcal{L}_{\mathrm{route}}=\frac{1}{|\Omega^s|}\sum_{\mathbf{x}\in\Omega^s}\mathrm{BCE}\big(g^s_R(\mathbf{x}),\ \sg(a^\ast(\mathbf{x}))\big),
\label{eq:lroute}
\end{equation}
where $\sigma$ is the sigmoid function, $T_a=0.25$ is a temperature and $\Omega^s$ is the set of object locations at stride $s$. The target depends only on the difference of the two losses: it is close to one where the RGB classifier has the lower loss and close to zero where the event classifier has the lower loss, so it measures the reliability of the events relative to the frame and not an absolute quality of either sensor. The loss is evaluated on object locations, where the choice of modality decides the detection. The target thus imposes at every object location the condition that QMF derives for whole samples, fusion weights that decrease with the loss of the corresponding unimodal predictor~\cite{zhang2023qmf}. The classifiers themselves are trained with the auxiliary loss $\mathcal{L}_{\mathrm{aux}}$, the sum of their focal losses normalized by the number of object locations. They are teachers used only in training: at inference, without labels, the router predicts the weights from the features alone, having learned to reproduce their judgment.

\textit{Object-level alignment.} The router compares two features, so the two modalities should describe the same object in a comparable way. For every ground-truth box $b$ of a batch $\mathcal{G}$, RoIAlign~\cite{he2017maskrcnn} extracts region features from $F^s_R$ and $F^s_E$ at the stride selected by the box size, and two projection heads map them to unit vectors $\mathbf{u}_R(b)$ and $\mathbf{u}_E(b)$. The symmetric InfoNCE loss~\cite{oord2018representation}
\begin{equation}
\mathcal{L}_{\mathrm{nce}}=-\frac{1}{2|\mathcal{G}|}\sum_{b\in\mathcal{G}}\sum_{(M,M')}\log\frac{e^{\langle\mathbf{u}_M(b),\mathbf{u}_{M'}(b)\rangle/\tau}}{\sum_{b'\in\mathcal{G}}e^{\langle\mathbf{u}_M(b),\mathbf{u}_{M'}(b')\rangle/\tau}},
\label{eq:nce}
\end{equation}
with $(M,M')\in\{(R,E),(E,R)\}$ and $\tau=0.1$, treats the two views of the same object as a positive pair and the other objects of the batch as negatives.

\subsection{Training Objective and Inference}
\label{sec:objective}
The total loss is
\begin{equation}
\mathcal{L}=\mathcal{L}_{\mathrm{det}}+\lambda_F\mathcal{L}_{\mathrm{F}}+\lambda_a\mathcal{L}_{\mathrm{aux}}+\lambda_r\mathcal{L}_{\mathrm{route}}+\lambda_c\mathcal{L}_{\mathrm{nce}},
\label{eq:total}
\end{equation}
where $\mathcal{L}_{\mathrm{det}}$ is the YOLOX detection loss with SimOTA assignment~\cite{ge2021yolox}, $\mathcal{L}_{\mathrm{F}}$ is the gate regularizer of FlexEvent, and $\lambda_F=10$, $\lambda_a=0.5$, $\lambda_r=0.5$ and $\lambda_c=0.1$. The auxiliary and routing losses are averaged over the three strides and the two event windows, and the alignment loss is computed on the primary window. The projection heads and all terms except $\mathcal{L}_{\mathrm{det}}$ are likewise used only during training. At inference, ERA-Det runs the event encoder on each new window, the RGB encoder on the latest frame, the router and the weighted sum at three strides, and the detection head.

\section{Experiments}
\label{sec:experiments}

\subsection{Dataset and Protocols}
\label{sec:data}
The DSEC-Det dataset~\cite{gehrig2024dagr} annotates 60 driving sequences of the DSEC dataset~\cite{gehrig2021dsec}, recorded with a Prophesee event camera at $640\times480$ and a color camera at 20~Hz whose frames are registered to the event camera. We follow the protocol of FlexEvent~\cite{lu2025flexevent}: 41 sequences for training, 6 for validation and 13 for testing, the two classes car and pedestrian, events downsampled to half resolution and cropped to $320\times215$, and boxes filtered by a minimum diagonal of 20 pixels and a minimum height of 10 pixels on the event grid. The frames are cropped to the same field of view, $640\times430$ for ERA-Det and $320\times215$ for FlexEvent. We report COCO average precision, with AP$_{50}$, AP$_{75}$, AP for small, medium and large objects, and AP per class.

Two evaluations are used. The Global evaluation detects at the annotated timestamps with 50~ms of events, the standard DSEC-Det setting. The interframe evaluation of DAGr and FlexEvent~\cite{gehrig2024dagr,lu2025flexevent} detects at $t=t_f+\Delta$ with $\Delta\in\{0, 5.6, 11.1, 16.7, 22.2, 27.8, 33.3, 50\}$~ms, which corresponds to operating frequencies of 180, 90, 60, 45, 36, 30 and 20~Hz at the fixed frame rate of 20~Hz and to the Empty setting without events at $\Delta=0$; the ground-truth boxes at $t$ are interpolated between consecutive annotations of each track. As a frequency-robustness measure we report the drop $\mathrm{AP}_{20}-\mathrm{AP}_{180}$ and the retention $\mathrm{AP}_{180}/\mathrm{AP}_{20}$.

\begin{table*}[!t]
\centering
\caption{Comparison with Event-Only and RGB--Event Detectors on the DSEC-Det Test Split, Two Classes. E: Events. F: Frames. $^\dagger$Results Reported in~\cite{lu2025flexevent}. FlexEvent + ARF Is ERA-Det without Reliability-Supervised Routing. Mean and Standard Deviation over Three Seeds for the Last Two Rows.}
\label{tab:global}
\renewcommand{\arraystretch}{1.12}%
\setlength{\tabcolsep}{5pt}
\resizebox{\textwidth}{!}{%
\begin{tabular}{@{}llcccccccc@{}}
\toprule
Input & Method & mAP & AP$_{50}$ & AP$_{75}$ & AP$_S$ & AP$_M$ & AP$_L$ & AP$_{\mathrm{car}}$ & AP$_{\mathrm{ped}}$ \\
\midrule
E & RVT$^\dagger$~\cite{gehrig2023rvt} & 38.4 & 58.7 & 41.3 & 29.5 & 50.3 & 81.7 & -- & -- \\
E & SAST$^\dagger$~\cite{peng2024sast} & 38.1 & 60.1 & 40.0 & 29.8 & 48.9 & 79.7 & -- & -- \\
E & SSM$^\dagger$~\cite{zubic2024ssm} & 38.0 & 55.2 & 40.6 & 28.8 & 52.2 & 77.8 & -- & -- \\
E & LEOD$^\dagger$~\cite{wu2024leod} & 41.1 & 65.2 & 43.6 & 35.1 & 47.3 & 73.3 & -- & -- \\
\midrule
E+F & DAGr-18$^\dagger$~\cite{gehrig2024dagr} & 37.6 & -- & -- & -- & -- & -- & -- & -- \\
E+F & DAGr-34$^\dagger$~\cite{gehrig2024dagr} & 39.0 & -- & -- & -- & -- & -- & -- & -- \\
E+F & DAGr-50$^\dagger$~\cite{gehrig2024dagr} & 41.9 & 66.0 & 44.3 & 36.3 & 56.2 & 77.8 & -- & -- \\
E+F & HDI-Former$^\dagger$~\cite{li2024hdiformer} & 46.7 & 69.1 & -- & -- & -- & -- & -- & -- \\
E+F & FlexEvent Balanced~\cite{lu2025flexevent} & 58.16 & 80.40 & 66.71 & 52.22 & 67.12 & \textbf{84.60} & -- & -- \\
E+F & FlexEvent Global~\cite{lu2025flexevent} & 59.53 & 79.25 & 69.81 & 54.57 & 68.86 & 83.23 & 72.96 & 46.11 \\
\midrule
E+F & FlexEvent + ARF & \textbf{63.41}$\pm$1.55 & 81.76$\pm$1.25 & \textbf{73.56}$\pm$1.72 & 58.72$\pm$0.91 & \textbf{71.25}$\pm$2.52 & 84.14$\pm$1.34 & 77.28$\pm$0.51 & \textbf{49.53}$\pm$2.90 \\
E+F & ERA-Det & 63.18$\pm$1.86 & \textbf{82.00}$\pm$1.47 & 73.28$\pm$2.43 & \textbf{58.89}$\pm$0.78 & 70.28$\pm$3.22 & 84.21$\pm$1.55 & \textbf{77.60}$\pm$0.27 & 48.75$\pm$3.44 \\
\bottomrule
\end{tabular}}
\end{table*}

\begin{table*}[!t]
\centering
\caption{Interframe Evaluation on the DSEC-Det Test Split. AP at Each Operating Frequency, Drop from 20 Hz to 180 Hz and Retention at 180 Hz. Empty Denotes Detection Without Events. FlexEvent + ARF Is ERA-Det without Reliability-Supervised Routing. Mean and Standard Deviation over Three Seeds for the Last Two Rows.}
\label{tab:interframe}
\setlength{\tabcolsep}{4pt}
\resizebox{\textwidth}{!}{%
\begin{tabular}{@{}lcccccccccc@{}}
\toprule
Method & Empty & 180 Hz & 90 Hz & 60 Hz & 45 Hz & 36 Hz & 30 Hz & 20 Hz & Drop & Retention \\
\midrule
FlexEvent Global~\cite{lu2025flexevent} & 34.20 & 46.98 & 55.60 & 57.69 & 59.10 & 60.17 & 60.46 & 60.12 & 13.14 & 78.1\% \\
FlexEvent Balanced~\cite{lu2025flexevent} & 37.40 & 52.64 & 56.77 & 58.40 & 59.32 & 59.66 & 59.62 & 59.06 & 6.42 & 89.1\% \\
FlexEvent + ARF & 51.51$\pm$3.92 & 58.07$\pm$1.76 & 63.00$\pm$1.40 & 64.41$\pm$1.55 & 65.11$\pm$1.52 & 65.52$\pm$1.51 & 65.41$\pm$1.53 & 64.37$\pm$1.58 & 6.29$\pm$0.65 & 90.2\% \\
ERA-Det &\textbf{55.55}$\pm$3.96 & \textbf{61.34}$\pm$1.84 & \textbf{64.24}$\pm$1.73 & \textbf{65.43}$\pm$1.86 & \textbf{65.77}$\pm$1.95 & \textbf{65.98}$\pm$1.87 & \textbf{65.68}$\pm$1.90 & \textbf{64.40}$\pm$1.89 & \textbf{3.07}$\pm$0.20 & \textbf{95.2\%} \\
\bottomrule
\end{tabular}}
\end{table*}

\begin{figure*}[!t]
\centering
\includegraphics[width=\textwidth]{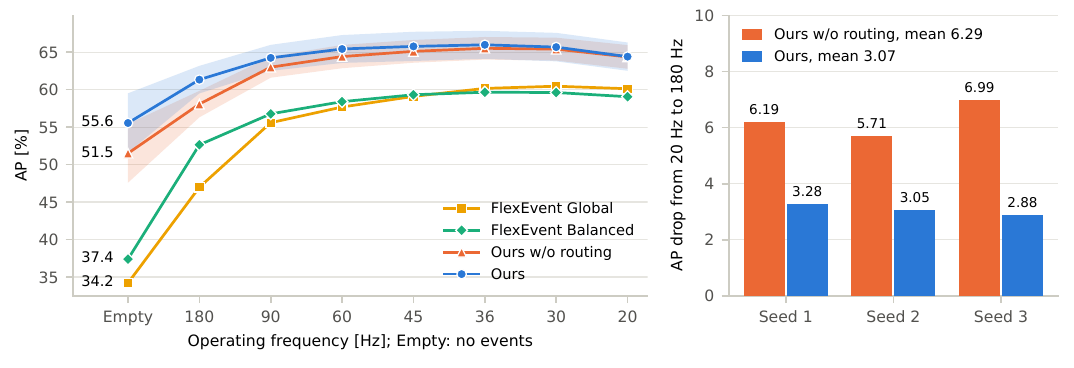}
\caption{Frequency robustness on the DSEC-Det test split. Left: AP from the Empty setting without events to 20~Hz; the shaded bands show one standard deviation over three seeds. Right: AP drop from 20~Hz to 180~Hz for each seed of ERA-Det and of the same detector without routing.}
\label{fig:frequency}
\end{figure*}

\begin{table*}[!t]
\centering
\caption{Global and Interframe Evaluation with the Eight-Class Protocol on the DSEC-Det Test Split: Pedestrian, Rider, Car, Bus, Truck, Bicycle and Motorcycle. Global Denotes mAP at the Annotated Timestamps. The Values of FlexEvent Balanced Are Those of the Official Release. FlexEvent + ARF Is ERA-Det without Reliability-Supervised Routing. Mean and Standard Deviation over Three Seeds for the Last Two Rows.}
\label{tab:eight}
\setlength{\tabcolsep}{4pt}
\resizebox{\textwidth}{!}{%
\begin{tabular}{@{}lccccccccccc@{}}
\toprule
Method & Global & Empty & 180 Hz & 90 Hz & 60 Hz & 45 Hz & 36 Hz & 30 Hz & 20 Hz & Drop & Retention \\
\midrule
FlexEvent Global~\cite{lu2025flexevent} & 53.40 & 27.87 & 41.85 & 47.70 & 49.68 & 51.95 & 52.95 & 54.29 & 54.62 & 12.77 & 76.6\% \\
FlexEvent Balanced~\cite{lu2025flexevent} & 51.76 & 33.99 & 45.57 & 50.82 & 52.66 & 53.18 & 53.86 & 54.19 & 53.38 & 7.81 & 85.4\% \\
FlexEvent + ARF & 55.12$\pm$0.99 & 38.23$\pm$2.80 & 48.58$\pm$0.38 & 52.94$\pm$1.17 & 53.90$\pm$1.35 & 55.28$\pm$1.30 & 55.68$\pm$1.37 & 56.64$\pm$1.12 & 56.26$\pm$0.94 & 7.68$\pm$0.59 & 86.3\% \\
ERA-Det & \textbf{55.71}$\pm$0.98 & \textbf{42.42}$\pm$3.42 & \textbf{52.01}$\pm$0.73 & \textbf{54.52}$\pm$0.57 & \textbf{55.16}$\pm$0.72 & \textbf{56.32}$\pm$0.56 & \textbf{56.45}$\pm$0.73 & \textbf{57.11}$\pm$0.83 & \textbf{56.90}$\pm$1.10 & \textbf{4.89}$\pm$0.62 & \textbf{91.4\%} \\
\bottomrule
\end{tabular}}
\end{table*}

\subsection{Baselines and Implementation Details}
\label{sec:impl}
FlexEvent is evaluated with its two released two-class checkpoints~\cite{lu2025flexevent}: Global, trained with the ground truth, and Balanced, which adds FlexTune pseudo-labels for high frequencies. Our evaluation of the Global checkpoint gives 59.53~mAP and 46.98~mAP at 180~Hz, against 59.52 and 46.94 in the official release, so the protocol reproduces the published numbers; the values of Balanced are taken from the official release. The released Global checkpoint exceeds the 57.4~mAP of the original paper and serves as the reference throughout. The event-only detectors RVT, SAST, SSM and LEOD and the RGB--event detectors DAGr and HDI-Former are compared with the results reported in~\cite{lu2025flexevent}, which retrained the event-only detectors on DSEC-Det and took DAGr and HDI-Former from their original papers. FlexEvent + ARF is fine-tuned with the schedule of ERA-Det.

ERA-Det is initialized from the Global checkpoint; the pooling layers, the router and the training heads are new. It is trained for 5,000 iterations with AdamW, a learning rate of $10^{-4}$ for the pretrained weights and $5\times10^{-4}$ for the new modules, a weight decay of $10^{-3}$, a cosine schedule with 3\% warm-up, gradient clipping at 1.0 and mixed precision. Each batch holds 16 sequences of 11 windows drawn by the mixed random and streaming sampling of FlexEvent, with horizontal flips and zoom augmentation. An exponential moving average of the weights with decay 0.9995 is evaluated on the validation split every 1,000 iterations, and the best validation checkpoint is evaluated once on the test split. Each model trains on one NVIDIA RTX 6000D GPU in about two hours. All results of ERA-Det and of its ablated variants in Tables~\ref{tab:global} to \ref{tab:components} are the mean and standard deviation over three training seeds; Table~\ref{tab:ablation} uses the first seed. Latency is measured at batch size 1 with half precision on the same GPU, over 300 test time steps after 30 warm-up steps, and includes both event windows, the RGB encoder, the neck, the head and non-maximum suppression.

\subsection{Comparison with State-of-the-Art Detectors}
\label{sec:comparison}
Table~\ref{tab:global} reports the Global evaluation. ERA-Det reaches 63.18~mAP and exceeds the strongest event-only detector, LEOD, by 22.1~mAP, the RGB--event detectors DAGr-50 and HDI-Former by 21.3 and 16.5~mAP, FlexEvent Global by 3.65~mAP and FlexEvent Balanced by 5.02~mAP. Its three seeds score 64.65, 61.10 and 63.78~mAP, so every trained model exceeds FlexEvent. Among object sizes the gain is largest for small objects, with AP$_S$ up by 4.32~mAP, and it is larger at AP$_{75}$, 3.47~mAP, than at AP$_{50}$, so the additional RGB pixels translate into more precise localization of small objects; the car class gains 4.64~mAP. FlexEvent + ARF reaches 63.41~mAP, within one standard deviation of ERA-Det, so the Global gain comes from asymmetric-resolution fusion, and routing contributes as the event window shortens.

Table~\ref{tab:interframe} and Fig.~\ref{fig:frequency} report the interframe evaluation. ERA-Det exceeds both FlexEvent models at every operating frequency, and its advantage grows as the event window shortens: 4.28~mAP over FlexEvent Global at 20~Hz, 14.36~mAP at 180~Hz and 21.35~mAP without events. Without pseudo-labels, ERA-Det also exceeds FlexEvent Balanced, which was trained with pseudo-labels for this regime, by 8.70~mAP at 180~Hz and by 18.15~mAP without events. Between 20~Hz and 180~Hz it loses 3.07~mAP and retains 95.2\% of its accuracy, against 78.1\% for FlexEvent Global and 89.1\% for FlexEvent Balanced, and between 20~Hz and 90~Hz it retains 99.8\%. At 180~Hz, ERA-Det, at 61.34~mAP, is more accurate than FlexEvent Global at 20~Hz, at 60.12~mAP.

\subsection{Eight-Class Protocol}
\label{sec:eight}
Table~\ref{tab:eight} repeats both evaluations with the eight-class protocol of DSEC-Det, which annotates pedestrians, riders, cars, buses, trucks, bicycles and motorcycles; the class train has no boxes and is not evaluated~\cite{lu2025flexevent}. FlexEvent + ARF and ERA-Det are fine-tuned from the released eight-class Global weights with the schedule of the two-class models, and our evaluation of the released model reproduces its published 53.40~mAP. ERA-Det reaches 55.71~mAP, 2.31~mAP above FlexEvent Global and 3.95~mAP above FlexEvent Balanced. At 180~Hz it reaches 52.01~mAP, 10.16~mAP above FlexEvent Global, 6.44~mAP above FlexEvent Balanced and 3.43~mAP above the same detector without routing, and without events it reaches 42.42~mAP, 14.55~mAP above FlexEvent Global. Routing lowers the drop from 20~Hz to 180~Hz from 7.68 to 4.89~mAP, against 12.77~mAP for FlexEvent Global, and lowers it in every seed. The two components thus act with seven classes as with two: asymmetric-resolution fusion raises the accuracy at every operating frequency, and reliability-supervised routing preserves it as the event window shortens.

\subsection{Contribution of Each Component}
\label{sec:components}
Table~\ref{tab:components} adds the components of ERA-Det to FlexEvent one at a time. All rows start from the same released weights.

\begin{table}[!t]
\centering
\caption{Contribution of Each Component on the DSEC-Det Test Split. ARF: Asymmetric-Resolution Fusion. RSR: Reliability-Supervised Routing. Mean and Standard Deviation over Three Seeds.}
\label{tab:components}
\setlength{\tabcolsep}{3pt}
\resizebox{\columnwidth}{!}{%
\begin{tabular}{@{}lccccc@{}}
\toprule
Configuration & Global & Empty & 180 Hz & 20 Hz & Drop \\
\midrule
FlexEvent Global & 59.53 & 34.20 & 46.98 & 60.12 & 13.14 \\
\quad + fine-tuning & 59.56$\pm$0.55 & -- & -- & -- & -- \\
\quad + ARF & 63.41$\pm$1.55 & 51.51$\pm$3.92 & 58.07$\pm$1.76 & 64.37$\pm$1.58 & 6.29$\pm$0.65 \\
\quad + ARF + RSR & 63.18$\pm$1.86 & \textbf{55.55}$\pm$3.96 & \textbf{61.34}$\pm$1.84 & 64.40$\pm$1.89 & \textbf{3.07}$\pm$0.20 \\
\bottomrule
\end{tabular}}
\end{table}

\textit{Fine-tuning control.} Fine-tuning FlexEvent at its original resolution for 8,000 iterations, with the same initialization, optimizer and exponential moving average, gives 59.56~mAP over three seeds, the accuracy of the released model. ERA-Det trains for 5,000 iterations, so its gains come from its components and not from additional training.

\textit{Asymmetric-resolution fusion.} Encoding the frame at its native resolution raises the Global accuracy by 3.88~mAP, to 63.41~mAP, and the accuracy at 180~Hz by 11.09~mAP, 2.6 times the gain of 4.25~mAP at 20~Hz: the frame carries a larger share of the evidence when the events are sparse, and a frame at native resolution provides more of it. The drop from 20~Hz to 180~Hz falls from 13.14 to 6.29~mAP.

\textit{Reliability-supervised routing.} Adding the router and its supervision keeps the Global and 20~Hz accuracy, raises the accuracy at 180~Hz by 3.27~mAP and without events by 4.04~mAP, and halves the drop from 20~Hz to 180~Hz, from 6.29 to 3.07~mAP. The effect holds in every seed, with drops of 3.28, 3.05 and 2.88~mAP against 6.19, 5.71 and 6.99~mAP without routing, as Fig.~\ref{fig:frequency} shows, and the standard deviation of the drop falls from 0.65 to 0.20~mAP. Routing therefore acts where the reliability of the two sensors changes, at short event windows.

\begin{figure*}[!t]
\centering
\includegraphics[width=\textwidth]{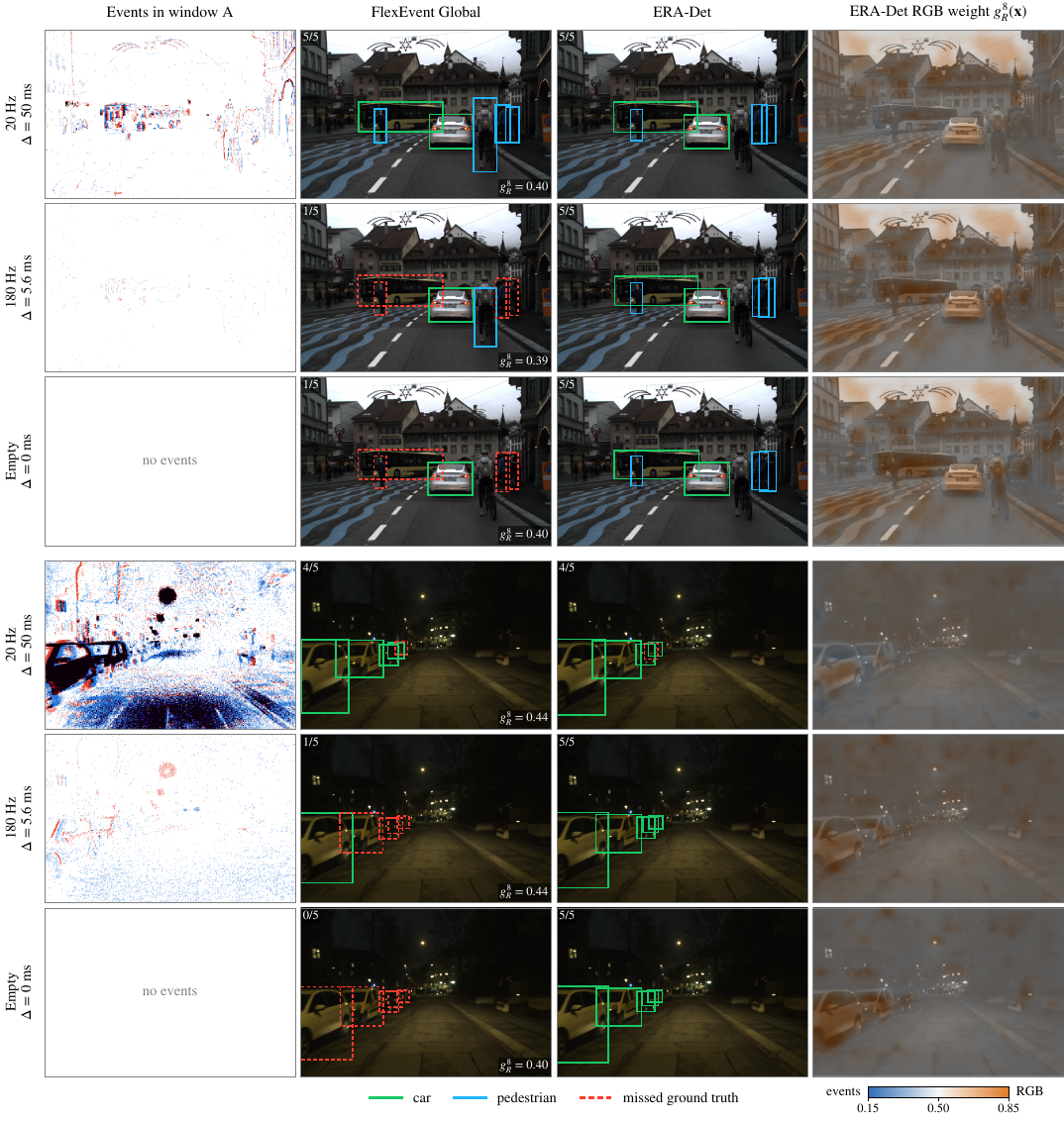}
\caption{Detections on two scenes of the DSEC-Det test split under the interframe protocol: a day scene of thun\_02\_a, top, and a night scene of zurich\_city\_12\_a, bottom. Each row pairs the same RGB frame with 50~ms of events, 20~Hz, with 5.6~ms of events, 180~Hz, or with no events, Empty. Columns, from left: events of the primary window, positive in red and negative in blue; detections of FlexEvent Global and of ERA-Det with a score of at least 0.3, cars in green and pedestrians in cyan, with missed ground-truth objects as red dashed boxes and the number of detected ground-truth objects at the upper left; the RGB weight $g^8_R(\mathbf{x})$ of ERA-Det at stride 8 over the frame, orange where the frame dominates and blue where the events dominate. The value at the lower right of the FlexEvent panels is its RGB weight at stride 8, shared by all locations.}
\label{fig:qualitative}
\end{figure*}

\subsection{Analysis of Reliability-Supervised Routing}
\label{sec:ablation}
Table~\ref{tab:ablation} removes one part of reliability-supervised routing at a time from the first seed of ERA-Det. The variants are compared by the drop from 20~Hz to 180~Hz, which is computed within each model and varies little across seeds, as shown in Table~\ref{tab:components}.

\begin{table}[!t]
\centering
\caption{Analysis of Reliability-Supervised Routing on the DSEC-Det Test Split, First Seed. All Variants Use Asymmetric-Resolution Fusion.}
\label{tab:ablation}
\setlength{\tabcolsep}{3pt}
\resizebox{\columnwidth}{!}{%
\begin{tabular}{@{}lcccccc@{}}
\toprule
Variant & Global & Empty & 180 Hz & 90 Hz & 20 Hz & Drop \\
\midrule
Full routing & 64.65 & 57.41 & 62.62 & 65.86 & 65.90 & 3.28 \\
Without per-location weights & 63.92 & 56.07 & 60.75 & 64.74 & 65.22 & 4.47 \\
Without routing supervision & 64.53 & 56.09 & 61.54 & 65.28 & 65.56 & 4.02 \\
Without statistics $\boldsymbol{\pi}^s$ & 64.79 & 57.78 & 62.86 & 66.05 & 65.94 & 3.08 \\
Without routing & 61.66 & 47.40 & 56.41 & 61.47 & 62.60 & 6.19 \\
\bottomrule
\end{tabular}}
\end{table}

\textit{Per-location weights.} Restricting the router to the per-sample gate of FlexEvent while keeping all training losses raises the drop from 3.28 to 4.47~mAP and lowers the accuracy at 180~Hz by 1.87~mAP. Reliability varies within a scene, and a weight per location captures this variation.

\textit{Routing supervision.} Training the per-location router with the detection loss alone, without the auxiliary classifiers, the routing target and the alignment loss, raises the drop to 4.02~mAP and lowers the accuracy without events by 1.32~mAP: the targets of Eq.~\ref{eq:target} provide the per-location signal that the fused detection loss lacks. The two parts are complementary: each recovers part of the robustness of the full design, and together they reduce the drop of the same detector without routing by 47\%.

\textit{Observation statistics.} Without the statistics $\boldsymbol{\pi}^s$, routing keeps its full effect, with a drop of 3.08 against 3.28~mAP, within the variation across seeds. Routing thus needs no hand-crafted prior: the supervised router learns the relative reliability of the two sensors from the features themselves, and the robustness of ERA-Det comes from the routing supervision.

\textit{Learned routing.} Averaged over the test split and the three seeds, the RGB weight $g^s_R$ is 0.53 at stride 8, 0.48 at stride 16 and 0.46 at stride 32, and the same order holds in each seed. The router gives the high-resolution frame more weight at the fine strides, where object detail is resolved, and the events more weight at the coarse strides, where motion context is aggregated, a division of labor that asymmetric-resolution fusion makes available.

\subsection{Routing Under Sparse Events}
\label{sec:routing}
Table~\ref{tab:routing} reports the mean RGB weight at each stride under the interframe protocol, over the 9,517 test frames of the protocol, for the first seed of ERA-Det and for FlexEvent Global. When the event window shortens from 50~ms to 5.6~ms, ERA-Det raises the RGB weight from 0.55 to 0.62 at stride 8 and from 0.49 to 0.55 at stride 16, and without events it holds 0.58 and 0.54. The shift is concentrated at the fine strides, and the weight at stride 32 changes by at most 0.02. The gate of FlexEvent moves by at most 0.03, and without events its weight at stride 8 falls from 0.41 to 0.40, toward the empty event stream. ERA-Det was trained with event windows of 50 and 25~ms only, so its response to windows of 5.6~ms and to the absence of events, learned from the reliability targets, transfers to observation conditions outside the training distribution.

\begin{table}[!t]
\centering
\caption{Mean RGB Weight $g^s_R$ per Stride $s$ under the Interframe Protocol on the DSEC-Det Test Split, First Seed of ERA-Det. FlexEvent Global Uses One Weight per Sample.}
\label{tab:routing}
\setlength{\tabcolsep}{6pt}
\begin{tabular}{@{}llccc@{}}
\toprule
Model & Setting & $s=8$ & $s=16$ & $s=32$ \\
\midrule
FlexEvent Global & 20 Hz & 0.41 & 0.45 & 0.48 \\
 & 180 Hz & 0.43 & 0.48 & 0.49 \\
 & Empty & 0.40 & 0.46 & 0.49 \\
\midrule
ERA-Det & 20 Hz & 0.55 & 0.49 & 0.46 \\
 & 180 Hz & \textbf{0.62} & \textbf{0.55} & 0.48 \\
 & Empty & 0.58 & 0.54 & 0.48 \\
\bottomrule
\end{tabular}
\end{table}

Fig.~\ref{fig:qualitative} shows two test scenes in which FlexEvent Global loses objects as the event window shortens. In the day scene, FlexEvent detects all five objects with 50~ms of events and one of them at 180~Hz and without events, while ERA-Det detects all five in every setting. In the night scene, FlexEvent detects four of five objects at 20~Hz, one at 180~Hz and none without events, while ERA-Det detects four, five and five. In the day scene, the moving bus produces dense events and is routed to the event stream at 20~Hz and to the frame when the events are absent. In the night scene, the weight on the parked cars, mixed at 20~Hz, turns toward the frame at 180~Hz and without events. The router thus redistributes the fusion location by location toward the sensor that observes each object, which a single weight per sample cannot express.

\begin{table}[!t]
\centering
\caption{Comparison on the PKU-DDD17-Car Test Split, Car Class. Day and Night: mAP on the 704 Day and 209 Night Test Samples. $^\dagger$Results Reported in~\cite{cao2024cafr}. FlexEvent + ARF Is ERA-Det without Reliability-Supervised Routing. The Last Three Rows Are Fine-Tuned from Their DSEC-Det Models; Mean and Standard Deviation over Three Seeds.}
\label{tab:ddd17}
\setlength{\tabcolsep}{4pt}
\resizebox{\columnwidth}{!}{%
\begin{tabular}{@{}lcccc@{}}
\toprule
Method & AP$_{50}$ & mAP & Day & Night \\
\midrule
SENet$^\dagger$~\cite{hu2018senet} & 81.6 & 42.4 & 43.7 & 37.0 \\
ECA-Net$^\dagger$~\cite{wang2020eca} & 82.2 & 40.8 & 42.2 & 36.1 \\
CBAM$^\dagger$~\cite{woo2018cbam} & 81.9 & 42.8 & 44.2 & 38.0 \\
SA-Gate$^\dagger$~\cite{chen2020sagate} & 82.0 & 43.4 & 44.9 & 38.0 \\
DCF$^\dagger$~\cite{ji2021dcf} & 83.4 & 42.5 & 43.4 & 39.0 \\
SPNet$^\dagger$~\cite{zhou2021spnet} & 84.7 & 43.3 & 44.9 & 37.1 \\
JDF$^\dagger$~\cite{li2019jdf} & 84.1 & -- & -- & -- \\
FPN-Fusion$^\dagger$~\cite{tomy2022fpnfusion} & 81.9 & 41.6 & 43.2 & 35.7 \\
DRFuser$^\dagger$~\cite{munir2023drfuser} & 82.6 & 42.4 & 43.3 & 38.8 \\
RAMNet$^\dagger$~\cite{gehrig2021ramnet} & 79.6 & 38.8 & 39.2 & 36.9 \\
CMX$^\dagger$~\cite{zhang2023cmx} & 80.4 & 39.0 & 40.2 & 35.4 \\
FAGC$^\dagger$~\cite{cao2021fagc} & 84.8 & 42.4 & 43.7 & 36.7 \\
RENet$^\dagger$~\cite{zhou2023renet} & 81.4 & 43.9 & 45.4 & 39.1 \\
EFNet$^\dagger$~\cite{sun2022efnet} & 83.0 & 41.6 & 43.4 & 35.1 \\
CAFR$^\dagger$~\cite{cao2024cafr} & 86.7 & 46.0 & 46.9 & 42.1 \\
\midrule
FlexEvent~\cite{lu2025flexevent} & 89.91$\pm$0.48 & 52.06$\pm$0.64 & 53.47$\pm$0.77 & \textbf{47.55}$\pm$0.68 \\
FlexEvent + ARF & \textbf{91.20}$\pm$0.83 & \textbf{52.99}$\pm$0.85 & \textbf{54.86}$\pm$1.07 & 47.15$\pm$0.23 \\
ERA-Det & 89.31$\pm$0.11 & 51.61$\pm$0.32 & 53.33$\pm$0.51 & 47.09$\pm$0.64 \\
\bottomrule
\end{tabular}}
\end{table}

\begin{table*}[!t]
\centering
\caption{Comparison on the PKU-DAVIS-SOD Test Split with Paired Frames and Events. E: Events. F: Frames. Normal, Low Light and Motion Blur: mAP per Test Condition. $^\dagger$Results Reported in~\cite{faod}. FlexEvent + ARF Is ERA-Det without Reliability-Supervised Routing. The Last Three Rows Are Fine-Tuned from Their DSEC-Det Eight-Class Models; Mean and Standard Deviation over Three Seeds.}
\label{tab:sod}
\setlength{\tabcolsep}{5pt}
\resizebox{\textwidth}{!}{%
\begin{tabular}{@{}llcccccccc@{}}
\toprule
Input & Method & mAP & AP$_{50}$ & AP$_{75}$ & AP$_S$ & AP$_M$ & Normal & Low light & Motion blur \\
\midrule
E & VIT-S5$^\dagger$~\cite{zubic2024ssm} & 23.2 & 46.6 & -- & -- & -- & -- & -- & -- \\
E & RVT$^\dagger$~\cite{gehrig2023rvt} & 25.6 & 50.3 & -- & -- & -- & -- & -- & -- \\
\midrule
F & YOLOX$^\dagger$~\cite{ge2021yolox} & 27.4 & 50.9 & -- & -- & -- & -- & -- & -- \\
F & VIT-S5$^\dagger$~\cite{zubic2024ssm} & 28.2 & 52.2 & -- & -- & -- & -- & -- & -- \\
F & RVT$^\dagger$~\cite{gehrig2023rvt} & 27.9 & 53.0 & -- & -- & -- & -- & -- & -- \\
\midrule
E+F & JDF$^\dagger$~\cite{li2019jdf} & -- & 44.2 & -- & -- & -- & -- & -- & -- \\
E+F & SODFormer$^\dagger$~\cite{li2023sodformer} & 20.7 & 50.4 & -- & -- & -- & -- & -- & -- \\
E+F & EFNet$^\dagger$~\cite{sun2022efnet} & 26.4 & 52.9 & -- & -- & -- & -- & -- & -- \\
E+F & AFNet$^\dagger$~\cite{zhang2023afnet} & 27.5 & 53.3 & -- & -- & -- & -- & -- & -- \\
E+F & FF-KDT$^\dagger$~\cite{wang2024ffkdt} & 28.5 & 54.2 & -- & -- & -- & -- & -- & -- \\
E+F & ReNet$^\dagger$~\cite{zhou2023renet} & 28.8 & 54.9 & -- & -- & -- & -- & -- & -- \\
E+F & FAOD$^\dagger$~\cite{faod} & \textbf{30.5} & 57.5 & -- & -- & -- & -- & -- & -- \\
\midrule
E+F & FlexEvent~\cite{lu2025flexevent} & 28.97$\pm$0.15 & 56.62$\pm$0.49 & 26.17$\pm$0.16 & 21.02$\pm$0.23 & 42.63$\pm$0.61 & 32.56$\pm$0.33 & 18.78$\pm$0.19 & 24.81$\pm$0.32 \\
E+F & FlexEvent + ARF & 29.85$\pm$0.13 & \textbf{58.52}$\pm$0.51 & 26.85$\pm$0.05 & 22.05$\pm$0.02 & \textbf{42.67}$\pm$0.31 & 33.55$\pm$0.12 & 19.60$\pm$0.12 & \textbf{25.69}$\pm$0.14 \\
E+F & ERA-Det & 29.77$\pm$0.06 & 58.17$\pm$0.03 & \textbf{27.03}$\pm$0.34 & \textbf{22.30}$\pm$0.10 & 42.11$\pm$0.51 & \textbf{33.62}$\pm$0.09 & \textbf{19.65}$\pm$0.37 & 25.33$\pm$0.18 \\
\bottomrule
\end{tabular}}
\end{table*}

\begin{table*}[!t]
\centering
\caption{Event Window and Frame Rate on the PKU-DAVIS-SOD Test Split, mAP. Event Window: the Frame of the Decision with the Last 50 to 0~ms of Events. Frame Rate: the Mismatch Protocol of FAOD, Frames at $25/N$~Hz and Events at 25~Hz; MDrop Is the Largest Loss Relative to $N=1$. $^\dagger$Results Reported in~\cite{faod}. First Seed for the Last Three Rows.}
\label{tab:sodrobust}
\setlength{\tabcolsep}{4pt}
\resizebox{\textwidth}{!}{%
\begin{tabular}{@{}lcccccc@{\hspace{10pt}}ccccccc@{}}
\toprule
 & \multicolumn{6}{c}{Event window} & \multicolumn{7}{c}{Frame rate $25/N$~Hz} \\
\cmidrule(lr){2-7}\cmidrule(l){8-14}
Method & 50 ms & 20 ms & 10 ms & 5 ms & 0 ms & Drop & $N=1$ & $N=2$ & $N=4$ & $N=6$ & $N=8$ & $N=10$ & MDrop \\
\midrule
SODFormer$^\dagger$~\cite{li2023sodformer} & -- & -- & -- & -- & -- & -- & 20.7 & 18.6 & 15.4 & 14.2 & 12.6 & 11.7 & 9.0 \\
EFNet$^\dagger$~\cite{sun2022efnet} & -- & -- & -- & -- & -- & -- & 26.4 & 24.8 & 21.6 & 19.8 & 16.7 & 15.9 & 10.5 \\
AFNet$^\dagger$~\cite{zhang2023afnet} & -- & -- & -- & -- & -- & -- & 27.5 & 25.3 & 21.2 & 20.0 & 16.5 & 15.1 & 12.4 \\
FF-KDT$^\dagger$~\cite{wang2024ffkdt} & -- & -- & -- & -- & -- & -- & 28.5 & 26.7 & 22.6 & 21.1 & 17.9 & 16.5 & 12.0 \\
ReNet$^\dagger$~\cite{zhou2023renet} & -- & -- & -- & -- & -- & -- & 28.8 & 27.0 & 22.8 & 21.4 & 18.3 & 17.0 & 11.8 \\
FAOD$^\dagger$~\cite{faod} & -- & -- & -- & -- & -- & -- & 29.7 & 29.5 & 29.2 & \textbf{29.1} & \textbf{28.6} & \textbf{28.5} & \textbf{1.2} \\
\midrule
FlexEvent~\cite{lu2025flexevent} & 28.85 & 27.84 & 24.94 & 22.14 & 16.89 & 11.96 & 28.85 & 29.02 & 28.83 & 28.35 & 27.72 & 27.24 & 1.61 \\
FlexEvent + ARF & \textbf{29.90} & \textbf{29.07} & 26.57 & 24.11 & 17.75 & 12.15 & \textbf{29.90} & \textbf{30.01} & 29.70 & 29.03 & 28.40 & 27.78 & 2.12 \\
ERA-Det & 29.85 & \textbf{29.07} & \textbf{27.00} & \textbf{25.10} & \textbf{19.78} & \textbf{10.07} & 29.85 & \textbf{30.01} & \textbf{29.72} & 29.04 & 28.46 & 27.82 & 2.03 \\
\bottomrule
\end{tabular}}
\end{table*}

\begin{table}[!t]
\centering
\caption{Mean RGB Weight $g^s_R$ at Object Locations per Stride $s$ on the PKU-DAVIS-SOD Test Split, Mean over Three Seeds. All: All Test Conditions. FlexEvent and FlexEvent + ARF Use One Weight per Sample.}
\label{tab:sodrouting}
\setlength{\tabcolsep}{6pt}
\begin{tabular}{@{}llccc@{}}
\toprule
Model & Setting & $s=8$ & $s=16$ & $s=32$ \\
\midrule
FlexEvent & Normal & 0.43 & 0.52 & 0.52 \\
 & Low light & 0.42 & 0.47 & 0.49 \\
 & Motion blur & 0.42 & 0.50 & 0.51 \\
 & All & 0.43 & 0.51 & 0.51 \\
 & All, no events & 0.49 & 0.57 & 0.55 \\
\midrule
FlexEvent + ARF & Normal & 0.46 & 0.51 & 0.51 \\
 & Low light & 0.44 & 0.45 & 0.50 \\
 & Motion blur & 0.45 & 0.49 & 0.51 \\
 & All & 0.45 & 0.50 & 0.51 \\
 & All, no events & 0.50 & 0.61 & 0.56 \\
\midrule
ERA-Det & Normal & 0.51 & 0.51 & 0.46 \\
 & Low light & \textbf{0.42} & \textbf{0.41} & \textbf{0.42} \\
 & Motion blur & 0.48 & 0.49 & 0.46 \\
 & All & 0.49 & 0.49 & 0.45 \\
 & All, no events & \textbf{0.69} & \textbf{0.72} & \textbf{0.59} \\
\bottomrule
\end{tabular}
\end{table}

\begin{figure*}[!t]
\centering
\includegraphics[width=\textwidth]{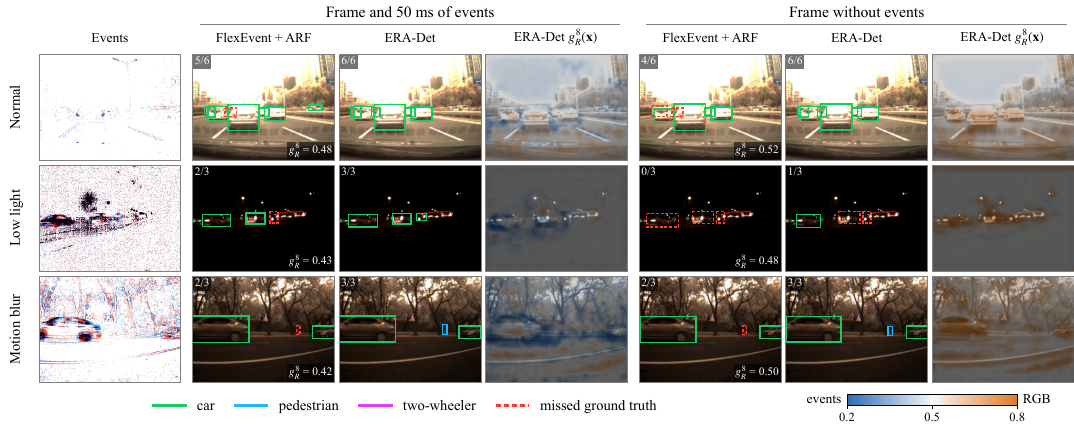}
\caption{Detections on three scenes of the PKU-DAVIS-SOD test split, one per test condition: normal, 005\_test\_normal, top; low light, 002\_test\_low\_light, middle; and motion blur, 009\_test\_motion\_blur, bottom. Each row pairs the same frame with the 50~ms of events that precede it, left, and with no events, right. Columns, from left: events of the window, positive in red and negative in blue; detections of FlexEvent + ARF and of ERA-Det with a score of at least 0.3, cars in green, pedestrians in cyan and two-wheelers in magenta, with missed ground-truth objects as red dashed boxes and the number of detected ground-truth objects at the upper left; the RGB weight $g^8_R(\mathbf{x})$ of ERA-Det at stride 8 over the frame, orange where the frame dominates and blue where the events dominate. The value at the lower right of the FlexEvent + ARF panels is its RGB weight at stride 8, shared by all locations.}
\label{fig:qualitative_sod}
\end{figure*}

\subsection{Transfer to Another Sensor}
\label{sec:ddd17}
The PKU-DDD17-Car dataset~\cite{li2019jdf} labels cars in the recordings of the DDD17 dataset~\cite{binas2017ddd17} of a DAVIS346 sensor, whose grey frames and events share one pixel array of $346\times260$ pixels, on highways and in cities, by day and at night. We follow the protocol of CAFR~\cite{cao2024cafr}: 2,241 samples for training and 913 for testing, of which 704 are recorded by day and 209 at night, and COCO AP of the car class with every box evaluated. Each sample holds a frame and the 20~ms of events that precede it, and consecutive samples are 1 to 2~s apart, so every sample is processed as a sequence of length one. FlexEvent, FlexEvent + ARF and ERA-Det are fine-tuned from their DSEC-Det models for 5,000 iterations with three seeds, and the final weights are evaluated because the protocol has no validation split. For the two detectors with asymmetric-resolution fusion, the frame is resized to twice the event resolution, so that their DSEC-Det weights apply unchanged, and the constants of the observation statistics are computed again on the training split.

Table~\ref{tab:ddd17} lists the results. Fine-tuned from DSEC-Det, all three detectors exceed every previously reported result, by 5.6 to 7.0~mAP over CAFR, so the fusion architecture and its DSEC-Det training transfer to a sensor with another resolution and grey frames. Asymmetric-resolution fusion raises FlexEvent to 52.99~mAP and 91.20~AP$_{50}$, the best results on this dataset, although the upsampled frame contributes no additional pixels: the gain comes from the larger scale at which the pretrained RGB encoder sees each car. ERA-Det lies within 1.4~mAP of the other two detectors. As every sample is a fixed 20~ms window that ends at its frame, the protocol has no interframe setting: each decision coincides with a new frame, so the gap between decision rate and frame rate that routing closes on DSEC-Det does not occur, and the frame carries most of the evidence. Without events each detector loses at most 1.4~mAP, against 25.9~mAP for FlexEvent Global on DSEC-Det.

The router follows the observation conditions on this sensor as well. Averaged over the three seeds and over object locations, the RGB weight at stride 16 is 0.58 by day and 0.50 at night, and it is lower at night at stride 32 as well: in the dark, where the frame loses contrast, the router moves the coarse strides toward the events. When the events are removed, it moves the weight toward the frame, from 0.54 to 0.64 at stride 8 and from 0.58 to 0.71 at stride 16 by day. Both orders hold in every seed, while the per-sample gate of FlexEvent stays between 0.50 and 0.54 in all conditions.

\subsection{Low Light, Motion Blur and Frame-Rate Mismatch}
\label{sec:sod}
The PKU-DAVIS-SOD dataset~\cite{li2023sodformer} records 220 driving sequences of about one minute with a color DAVIS346 sensor, whose 25~Hz frames and events share an array of $346\times260$ pixels, in normal, low-light and motion-blur conditions, and labels cars, pedestrians and two-wheelers in every frame. We use the release and the protocol of FAOD~\cite{faod}: 132 recordings for training and 44 for testing, of which 28 are normal, 6 low light and 10 motion blur; boxes are clipped to the image and kept when they are at least 5 pixels wide and high; each frame is paired with the 50~ms of events that precede it; and the first 0.5~s of each recording is not evaluated, which leaves 52,543 test frames. FlexEvent, FlexEvent + ARF and ERA-Det are fine-tuned from their DSEC-Det eight-class models for 5,000 iterations with three seeds, on streams of 11 frames as on DSEC-Det, and the final weights are evaluated. The class weights for car and pedestrian are copied from the same DSEC-Det classes and those for two-wheeler from the mean of bicycle and motorcycle. For the two detectors with asymmetric-resolution fusion the frame is resized to twice the event resolution, and the constants of the observation statistics are computed again on the training split. In half of the training windows, the RGB input of all three detectors is a frame one to four steps older than the decision.

Table~\ref{tab:sod} compares the three detectors with the results reported under the same protocol~\cite{faod}. FlexEvent + ARF and ERA-Det reach 58.52 and 58.17~AP$_{50}$, above every reported detector and 1.0 and 0.7~AP$_{50}$ above FAOD, and 29.85 and 29.77~mAP, above every reported detector except FAOD at 30.5~mAP. Asymmetric-resolution fusion raises FlexEvent by 0.88~mAP and 1.90~AP$_{50}$ with an upsampled frame, as on PKU-DDD17-Car, and in each test condition. With paired frames and events, ERA-Det and FlexEvent + ARF lie within 0.08~mAP, as in the Global evaluation of DSEC-Det, and ERA-Det has the highest AP$_{75}$ and AP$_S$ of the three detectors.

Table~\ref{tab:sodrobust} shortens the event window before the frame. From 50 to 5~ms of events, ERA-Det loses 4.75~mAP, FlexEvent + ARF 5.79~mAP and FlexEvent 6.71~mAP, and without events ERA-Det loses 10.07~mAP against 12.15 and 11.96~mAP. Over three seeds, ERA-Det reaches 19.47~mAP without events, 1.48~mAP above FlexEvent + ARF and 2.59~mAP above FlexEvent, and exceeds FlexEvent + ARF in every seed. Without events every detector loses 10 to 12~mAP on this dataset, against at most 1.4~mAP on PKU-DDD17-Car, so the events carry a large share of the evidence and routing preserves the accuracy as they become sparse, as on DSEC-Det.

The frame-rate protocol of FAOD keeps the events at 25~Hz and delivers the frames at $25/N$~Hz, so each decision uses the latest frame, which is up to $N-1$ steps old. Unlike the operating frequency on DSEC-Det, which raises the decision rate at a fixed frame rate, this protocol lowers the frame rate at a fixed decision rate. The reported fusion detectors lose 9.0 to 12.4~mAP at $N=10$, and FAOD, which aligns the RGB features to the events, loses 1.2~mAP. The three detectors lose 1.61 to 2.12~mAP and reach 27.24 to 27.82~mAP at $N=10$, 10.8~mAP above ReNet. All three detectors see frames up to four steps older than the decision during training, a delay that the protocol reaches at $N=5$.

Table~\ref{tab:sodrouting} reports the RGB weight at object locations. In normal scenes ERA-Det gives the frame a weight of 0.51 at strides 8 and 16, in low light it lowers the weight to 0.42 and 0.41, and under motion blur to 0.48 and 0.49. The decrease from normal scenes to low light holds in every seed and is at least 0.08 at both strides: where the frame loses contrast, the router moves the fusion toward the events. The per-sample gate of FlexEvent + ARF falls by 0.02 and 0.06 at the same strides and that of FlexEvent by 0.01 and 0.05. Without events, ERA-Det raises the RGB weight at stride 8 from 0.49 to 0.69, against 0.45 to 0.50 for FlexEvent + ARF and 0.43 to 0.49 for FlexEvent. The router thus follows the reliability of the frame in both directions, toward the events when the frame degrades and toward the frame when the events vanish.

Fig.~\ref{fig:qualitative_sod} shows one test scene per condition. With events, ERA-Det detects all six cars of the normal scene, all three cars in low light and all three objects under motion blur, including a pedestrian that FlexEvent + ARF misses, while FlexEvent + ARF detects five, two and two of them. Without events, ERA-Det detects six, one and three objects and FlexEvent + ARF four, none and two. On the objects, the RGB weight of ERA-Det is 0.51 in the normal scene, 0.42 in low light and 0.36 under motion blur, where the moving car produces dense events, and it rises to 0.72, 0.70 and 0.70 without events, while the single weight of FlexEvent + ARF stays between 0.42 and 0.52.

\subsection{Efficiency}
\label{sec:efficiency}
Table~\ref{tab:efficiency} reports the cost per time step, with the frame encoded anew at every time step. Asymmetric-resolution fusion raises the computation from 36 to 95~GFLOPs, all of it in the RGB encoder, and the GPU latency by 0.1~ms: at batch size 1 the latency is set by the number of sequential operations, and the larger RGB convolutions use GPU capacity that the recurrent event encoder leaves idle. Reliability-supervised routing adds 0.05~M parameters and 2.8~ms, including the observation statistics, and the training heads add no cost at inference, so ERA-Det runs in 22.0~ms per time step, 2.9~ms more than FlexEvent. Because the RGB features depend only on the frame, a deployment computes them once per frame and reuses them for all decisions until the next frame arrives. At 180~Hz, nine decisions share one RGB encoding, which divides the additional computation per decision by nine.

\begin{table}[!t]
\centering
\caption{Cost per Time Step at Batch Size 1 and Half Precision on an NVIDIA RTX 6000D GPU, with Global mAP.}
\label{tab:efficiency}
\setlength{\tabcolsep}{4.2pt}
\begin{tabular}{@{}lcccc@{}}
\toprule
Model & Params/M & GFLOPs & Latency/ms & mAP \\
\midrule
FlexEvent Global & 45.59 & 36 & 19.1 & 59.53 \\
\quad + ARF & 45.59 & 95 & 19.2 & 63.41 \\
ERA-Det & 45.64 & 95 & 22.0 & 63.18 \\
\bottomrule
\end{tabular}
\end{table}

\section{Conclusion}
\label{sec:conclusion}
This paper presented ERA-Det, an RGB--event detector that stays accurate as its operating frequency, the rate of its decisions, rises far above the frame rate. Its principle is that the fusion should follow, location by location, the reliability of the events relative to the frame. Reliability-supervised routing realizes it by supervising a fusion weight at every location with targets from two single-modality classifiers, a supervision that earlier fusion methods apply only to whole samples, and asymmetric-resolution fusion gives the frame the evidence that routing relies on by decoupling its resolution from that of the events while every pretrained weight is reused. ERA-Det sets a new state of the art on DSEC-Det at 63.2~mAP, 3.7~mAP above FlexEvent, and is the most accurate detector at every operating frequency from 20~Hz to 180~Hz: at 180~Hz it reaches 61.3~mAP, 8.7~mAP above the pseudo-label variant of FlexEvent and above the accuracy of FlexEvent at 20~Hz. Routing halves the accuracy loss between 20~Hz and 180~Hz and lowers it in every seed. Both components remain effective with the seven annotated classes of DSEC-Det, and on two DAVIS346 datasets asymmetric-resolution fusion raises FlexEvent by 0.9~mAP although the frame has the resolution of the events. The router moves the fusion toward the frame when the events become sparse and toward the events at night and in low light, and without events it adds 1.5~mAP on PKU-DAVIS-SOD. ERA-Det is obtained by fine-tuning released weights for only 5,000 iterations on a single GPU. Asymmetric-resolution fusion applies to any fusion detector that encodes the frame at the event resolution, and reliability-supervised routing raises the accuracy wherever the reliability of the events relative to the frame changes within a sequence, as between frames at high decision rates. Such non-uniform reliability is shared by other tasks that combine events with frames, such as event-based SLAM~\cite{vidal2018ultimate}, and future work will carry routing to them, to other sensor pairs such as RGB and thermal or LiDAR, and to embedded platforms where the decision rate determines the reaction time of the vehicle.

\bibliographystyle{IEEEtran}
\bibliography{cas-refs}

\end{document}